\documentclass[a4paper, 10pt, conference]{ieeeconf}      

\IEEEoverridecommandlockouts                              
\usepackage{graphics} 
\usepackage{graphicx}
\usepackage{mathptmx} 
\usepackage{times} 
\usepackage{amsmath} 
\usepackage{amssymb}  
\usepackage{color}

\usepackage{float}
\usepackage{stfloats}

\title{\LARGE \bf
\emph{Hoxels}: Fully 3-D Printed Soft Multi-Modal \& Multi-Contact \\Haptic Voxel Displays for Enriched Tactile Information Transfer 
}

\author{Zhenishbek Zhakypov, Yimeng Qin, and Allison M.~Okamura
\thanks{*This work is funded in part by the Swiss National Science Foundation (SNSF) Early Postdoc Mobility grant P2ELP2$\_$195132 and US National Science Foundation grants 1830163 and 1812966.}
\thanks{All authors are with the department of Mechanical Engineering, Stanford University, Stanford, CA 94301, USA.
        {\tt\small zhakypov@stanford.edu, yimengq@stanford.edu, aokamura@stanford.edu}}%
}

\begin{document}

\maketitle
\thispagestyle{empty}
\pagestyle{empty}

\begin{abstract}
Wrist-worn haptic interfaces can deliver a wide range of tactile cues for communication of information and interaction with virtual objects. Unlike fingertips, the wrist and forearm provide a considerably large area of skin that allows placement of multiple haptic actuators as a display for enriching tactile information transfer with minimal encumbrance. Existing multi-degree-of-freedom (DoF) wrist-worn devices employ traditional rigid robotic mechanisms and electric motors that limit their versatility, miniaturization, distribution, and assembly. Alternative solutions based on soft elastomeric actuator arrays constitute only 1-DoF haptic pixels. Higher-DoF prototypes produce a single interaction point and require complex manual assembly processes, such as molding and gluing several parts. These approaches limit the construction of high-DoF compact haptic displays, repeatability, and customizability. Here we present a novel, fully 3D-printed, soft, wearable haptic display for increasing tactile information transfer on the wrist and forearm with 3-DoF haptic voxels, called \emph{hoxels}. Our initial prototype comprises two hoxels that provide skin shear, pressure, twist, stretch, squeeze, and other arbitrary stimuli. Each hoxel generates force up to 1.6 N in the $x$ and $y$-axes and up to 20 N  in the $z$-axis. Our method enables the rapid fabrication of versatile and forceful haptic displays.
\end{abstract}

\section{INTRODUCTION}

\begin{figure}[t]
\centering
\includegraphics[width=\columnwidth]{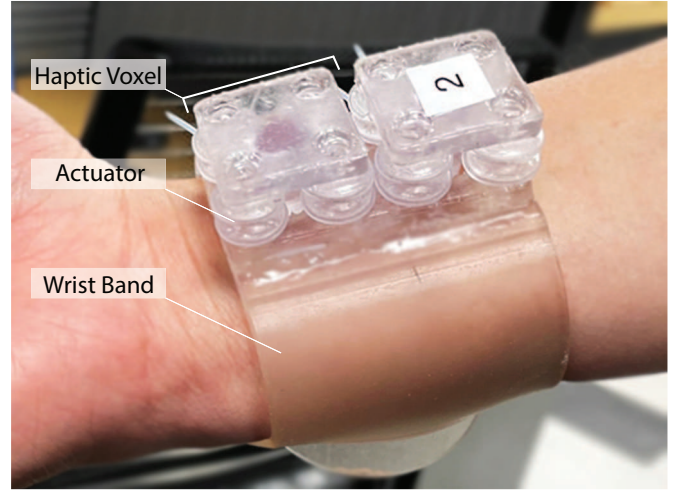}
\caption{The novel, fully 3D-printed, soft wearable haptic voxel (\emph{hoxel}) display for distributed and rich skin tactile stimulation on the wrist (volar side). Each haptic voxel comprises four bellow type actuators that when actuated selectively move the tactor pin at the center in 3-DoF: $x$ and $y$ for skin lateral shear and $z$ for pressure. Multiple hoxels can stimulate a variety of skin interactions, such as rotational shear or twist, stretch, squeeze, and other tactile stimulations.}
\label{fig:prototype}
\end{figure}

Besides vision, the tactile sensation is critical for the human sensorimotor experience. Skin deformations due to touch, including shear,  pressure, stretch, squeeze, or vibrations can reveal various physical attributes of contact~\cite{pezent2019tasbi}. Wrist-worn robotic haptic devices and displays can substitute for complex skin interactions on the wrist and forearm. Unlike fingertip haptic devices, they expose the fingers and do not face common limitations, such as physical interference between multi-finger devices when grasping small-sized virtual objects and finger tracking with virtual reality (VR) headset cameras and infrared sensors. Additionally, the large skin area on the wrist and forearm (both dorsal and volar sides) provides a convenient space and possibility to interface with multi-actuator haptic displays for increased tactile information transfer, potentially improving communication and immersion in VR. 

Building a versatile and forceful haptic display is challenging. First, each actuator unit in a haptic display should generate multi-DoF motions and forces to cause complex skin deformations with single contact. Second, they should be compact to integrate them into scalable displays for multi-contact interactions. Third, a haptic display should be simple to fabricate and interface with the human body. Creating these abilities with traditional robotic technology is infeasible due to trade-offs among multifunctionality, miniaturization, and manufacturability (which we call the ``3M problem''). Traditional link-pin-joint multi-DoF mechanisms and electric motors are bulky and heavy, and difficult to assemble and expensive. Many existing prototypes reduce complexity to deliver limited haptic cues, like vibrations~\cite{hong2017vibro}, 1-DoF motion for skin lateral shear~\cite{chinello2016shear}, rotational shear~\cite{bark2010rotational}, and 2-DoF motion for shear and pressure~\cite{SaracRAL2022, moriyama2018stretchpressure}. It requires new interdisciplinary design methodologies, materials, and structures. 

Soft and smart materials offer a promising method for designing multifunctional wearable haptic devices. Researchers presented various soft actuator arrays based on vibrotactile fluidic~\cite{sonar2016pneumatic}, shape memory polymers~\cite{besse2017smp} and dielectric elastomer actuators (DEA)~\cite{zhao2020dea}, also referred to as haptic displays or artificial skins. Although they increase the amount of tactile information by distributing several actuators or haptic pixels on the skin surface, each pixel can provide only a 1-DoF pressure stimulus. Recently, more capable soft haptic devices have been developed that can produce skin deformations with up to 3-DoF tactor movements~\cite{YoshidaWHC2019}. However, these prototypes generate only a single contact point and are bulky for constituting haptic displays. They also require complex manual assembly processes, such as molding and gluing several parts. Creating high-DoF mechanisms, scalability, design repeatability, and customizability remains a considerable challenge in the soft robotics field. 

Here we present a monolithically 3-D printed soft haptic voxel display, which we call a \emph{hoxel}, for simulating complex and distributed tactile interactions on the wrist and forearm (Fig.~\ref{fig:prototype}). Each hoxel comprises four bellow type actuators that when vacuum-powered selectively move the tactor in 3-DoF: $x$ and $y$ for skin lateral shear and $z$ for pressure. A combination of multiple hoxels can stimulate diverse skin deformations, such as lateral shear, twist, stretch, and squeeze, and many other. We present the first prototype with two hoxels and blocked force test results. Each hoxel generates forces up to 1.6 N in $x$ and $y$-axes and up to 20 N  in $z$-axis, demonstrating versatility, scalability, and high-power actuation.

\begin{figure}[t]
\centering
\includegraphics[width=\columnwidth]{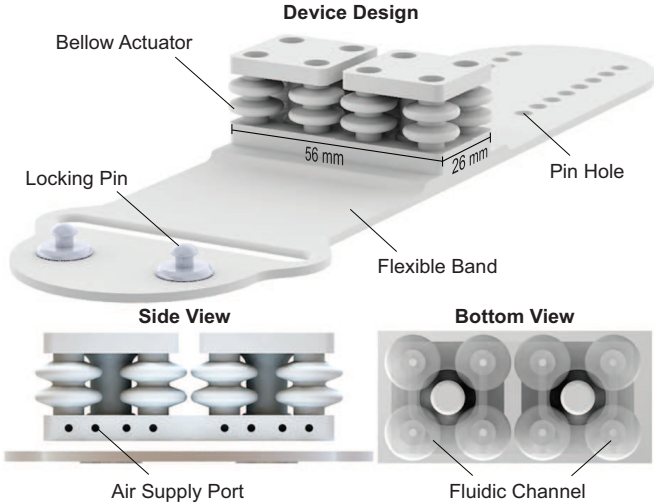}
\caption{The design of the proposed soft haptic display with two hoxel actuators and a flexible wristband. Each hoxel comprises four bellow type vacuum actuators with a tactor pin at the center. The vacuum air is supplied independently to the actuators via eight air supply ports and embedded fluidic channels. The adjustable flexible wristband with locking pins allows for interfacing the device with the wrist.}
\label{fig:design}
\end{figure}

\section{Device Design \& Fabrication}  

\begin{figure*}[t]
\centering
\includegraphics[width=\textwidth]{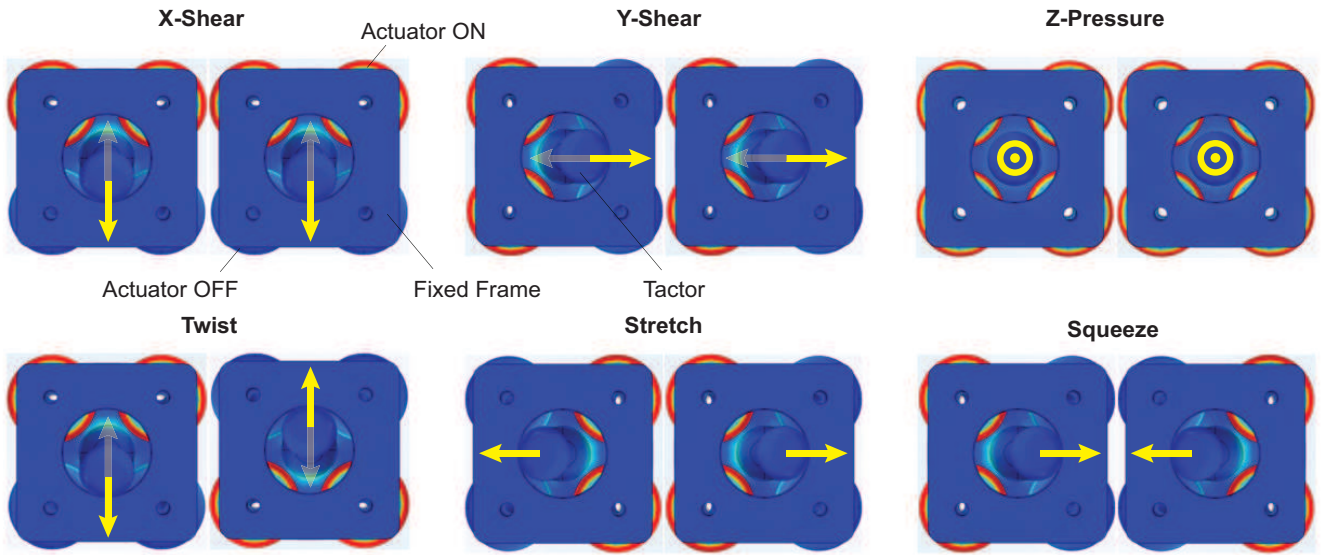}
\caption{Two hoxels can deliver diverse skin deformation stimuli: skin $x$ and $y$ shear, $z$ pressure, twist, stretch, and squeeze. Although not illustrated here, the device can produce other arbitrary haptic stimuli owing to the tactors' freedom of motion.}
\label{fig:movements}
\end{figure*}

Conventional haptic actuator displays distribute multiple one-DoF tactile interactions over the skin surface. Each haptic actuator conveys only one-dimensional tactile information, like a pixel in a digital screen. We achieve three-dimensional information transfer with hoxels (Fig.~\ref{fig:prototype} and ~\ref{fig:design}). Each hoxel produces 3-DoF rotational tactor movement to cause skin shear and pressure. When combined in an array they generate several other complex interactions, such as skin twist, stretch, squeeze, and others, as in Fig.~\ref{fig:movements}. Four vacuum-type bellow actuator columns placed in parallel compress individually to move and rotate the square platform with tactor. As a result, the tactor produces roll and yaw motions for skin shear in $x$ and $z$-axes, respectively, and a linear motion in $z$-axis for pressure. The display also houses fluidic channels to distribute air supply to the actuators.            

To interface the haptic display with the wrist, we design a flexible wristband with pin-hole locking mechanism as in Fig.~\ref{fig:design}, similar to those used in smart watches. The band is wide enough to ensure user comfort by distributing and minimizing reaction forces arising from attachment and actuation.

To enable rapid fabrication and scalability of the proposed design with minimal assembly effort, we 3-D printed the device monolithically using a commercial Stereolithography 3-D printer (Form 3, Formlabs Inc.) and soft material resin (Flexible 80A, Formlabs Inc.). We extend our earlier fabrication method of a fully 3-D printed 4-DoF fingertip device~\cite{ZhakypovROBOSOFT2022} to the printing of the wrist-worn haptic display. The model preparation and post-processing are similar and the details can be found in~\cite{ZhakypovROBOSOFT2022}. Although our approach allows for 3-D printing the entire design monolithically, we 3-D print the hoxel in one piece and the wristband separately in two pieces  owing to size limitation of the print bed. We bond the wristband to the haptic display using the same resin as the final touch. 

\begin{figure}[t]
\centering
\includegraphics[width=\columnwidth]{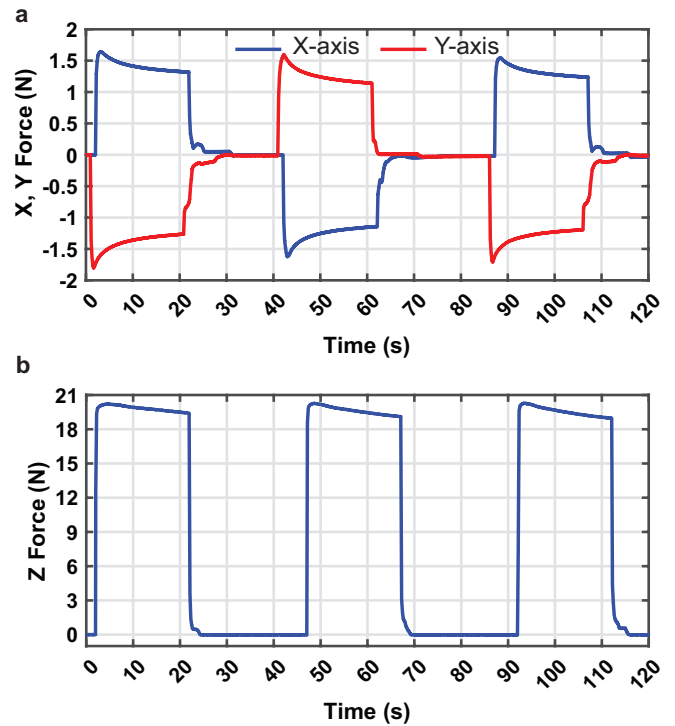}
\caption{Blocked force measurement results for single haptic voxel. The tactor produces approximately (a) 1.6 N in the $x$ and $y$ axes (parallel to the skin) and (b) 20 N in the $z$ axis (normal to the skin). The corresponding actuators were activated and deactivated with periods of 20 sec. The soft tactor bends when actuated under the blocked condition and the material viscoelasticity causes force relaxation.}
\label{fig:forces}
\end{figure}

\section{Force Characterization}

To evaluate the force capacity of each hoxel, we measured one tactor's blocked force for all three DoFs. We employed a Nano-17 force sensor (ATI Industrial Automation, Inc.) and attached the device to the sensor via a 3D-printed rigid adapter. The experimental results for the blocked force test are presented in Fig.~\ref{fig:forces}. The force magnitude for the $x$ and $y$ axes are similar (Fig.~\ref{fig:forces}(a)) at around $\pm$1.6 N for maximum applied vacuum ($\approx$ -100 KPa) and the maximum $z$-axis force was considerably larger at approximately 20 N (see Fig.~\ref{fig:forces}(b). The force output of the device is suitable for haptic stimulation, particularly because humans are more sensitive to stimulation in the shear ($x$ and $y$) directions than in the normal ($z$) direction.

\section{Conclusion}

We developed a novel, soft, fully 3-D printable haptic display based on haptic voxels (hoxels) that produce 3-DoF motions and forces, and when combined in an array can deliver a variety of tactile stimuli on the wrist. The proposed method enables rapid manufacturing of miniature yet distributed mechanism arrays and their interfacing with user's body with minimal assembly. We demonstrated the effectiveness of the method for building miniature haptic interfaces with adequate force output. We plan to study diverse interaction and grasping scenarios in virtual environments by localizing the hoxels on the wrist's dorsal and volar sides. 








\bibliographystyle{IEEEtran}
\bibliography{CHARMBib}

\end{document}